\documentclass[letterpaper, 10 pt, conference]{ieeeconf}  

\IEEEoverridecommandlockouts                              

\usepackage{cite}
\usepackage{amsmath,amssymb,amsfonts}
\usepackage{algorithmic,algorithm}
\usepackage{graphicx}
\usepackage{textcomp}
\usepackage{xcolor}
\usepackage{caption}
\usepackage{subcaption} 

\usepackage{lipsum}
\newtheorem{definition}{Definition}
\newtheorem{theorem}{Theorem}[section]

\newtheorem{lemma}[theorem]{Lemma}

\def\BibTeX{{\rm B\kern-.05em{\sc i\kern-.025em b}\kern-.08em
    T\kern-.1667em\lower.7ex\hbox{E}\kern-.125emX}}

\title{\LARGE \bf
Group-based control of large-scale micro-robot swarms with on-board Physical Finite-State Machines}
\author{Siyu Li, Milo\v s \v Zefran, and Igor Paprotny
\thanks{All authors are with the Department of Electrical and Computer Engineering, University of Illinois Chicago, Chicago, IL 60607, USA. Email: {\tt \{sli230,mzefran,paprotny\}@uic.edu}}%
\thanks{ This work has been supported by the National Science Foundation grants IIS-1705058 and CMMI-1762924.}%
}

\newcommand{\comment}[1]{}

\usepackage{mathtools}

\begin{document}
\maketitle
\thispagestyle{empty}
\pagestyle{empty}

\begin{abstract}
An important problem in microrobotics is how to control a large group of microrobots with a global control signal. This paper focuses on controlling a large-scale swarm of MicroStressBots with on-board physical finite-state machines. We introduce the concept of group-based control, which makes it possible to scale up the swarm size while reducing the complexity both of robot fabrication as well as swarm control. We prove that the group-based control system is locally accessible in terms of the robot positions. We further hypothesize based on extensive simulations that the system is globally controllable. A nonlinear optimization strategy is proposed to control the swarm by minimizing control effort. We also propose a probabilistically complete collision avoidance method that is suitable for online use. The paper concludes with an evaluation of the proposed methods in simulations.
\end{abstract}

\section{Introduction}
Micro-robot systems have promising applications in medicine~\cite{chowdhury_controlling_2015,li_development_2018,yu_novel_2010,leclerc_vitro_2020,mahdy_ultrasound-guided_2018} and micro-assembly~\cite{diller_three-dimensional_2014,floyd_two-dimensional_2009,yao_directed_2020}.
There are several realizations of the micro-robot systems.
Robots can be actuated using electrostatic fields, magnetic fields, physiological energy, or optical energy~\cite{banerjee2012real,pawashe2009modeling}. This paper focuses on the stress-engineered MEMS micro-robots (MicroStressBot), originally developed in~\cite{donald_untethered_2006}, which are powered by a uniform (electrostatic) field generated by a substrate underneath the robots. It is important to notice that all these micro-robot systems are controlled by applying a global signal to all the robots within the operating environment. This is because the size restricts robots from enough on-board control logic and power storage. Thus, achieving a task such as micro-assembly or drug delivery with micro-scale robots requires the coordination of a large group of micro-robots simultaneously.

At present, parallel control of multiple micro-robots relies on robots being designed to be physically different in order to elicit different behavior in response to a single central control signal. Such different behavior can in turn be exploited to control the robot swarm as a whole. Ensemble Control (EC) was proposed in~\cite{becker2014controlling}. It assumes that robots have different linear velocity and turning rate parameters and asymptotically stable behavior is obtained by using $k$-step pseudo-inverse control. Turning-rate Selective Control (TSC)~\cite{paprotny_turning-rate_2013} achieves individual control by explicitly designing robots to have different turning rates. These approaches are difficult to scale to larger swarms. In~\cite{yu2018pattern}, a rotating magnetic field is used to generate a vortex-like paramagnetic nanoparticle swarm (VPN), which will change according to the input rotating frequency.

In~\cite{donald2013planning}, Global Control Selective Response (GCSR) approach was proposed that relies on the separation between levels of the control signal to elicit differences in motion. Using MicroStressBots, it introduced a 2-state Physical finite-state machine (FSM) which allows robots to be controlled by using 2 voltage levels. It has been shown that $n$ robots can be controlled using $O(\sqrt{n})$ voltage levels. Subsequently, \cite{paprotny2017finite} showed that only $O(1)$, a constant number of voltage levels suffice to control a swarm of $n$ robots if they are equipped with on-board multi-stage Physical FSM (PFSM). Implicitly, the assumption in~\cite{paprotny2017finite} was that the robots can be individually addressed and activated one by one. In this paper, we take this idea further and introduce \emph{group-based control}. We show that the approach both dramatically reduces the size of the PFSMs needed for the implementation as well as greatly reduces the time needed to move the swarm from the initial to the final configuration.

Another aspect that our paper considers is collision avoidance. In~\cite{becker2014controlling}, collision avoidance has been studied in the probabilistic sense, but no explicit collision avoidance algorithm has been proposed. Sample-based methods like RRT~\cite{kuffner_rrt-connect_2000} and its variants are commonly used in the robotics community for motion planning subject to constraints such as collision avoidance. In~\cite{vcap2013multi}, a multi-agent RRT$^*$ is evaluated and shown to have good scalability in a large sparse environment. In~\cite{chiang2019rl}, a reinforcement learning method is proposed to navigate the robot to a goal through an efficient path. Recent work~\cite{haghtalab2018provable} has shown how these methods can be made more efficient. These approaches all assume that each robot is individually controlled and are not directly applicable when a robot swarm is controlled with a single global control input.
In our case, random walks are employed to allow the robots to avoid the collision, similarly to~\cite{barraquand1991robot,carpin2005motion}, but the approach is adapted to the case of a single global control signal.

\section{Background}

\subsection{MicroStressBot}

Electrostatic stress-engineered MEMS microrobot (MicroStressBot) is a 120 $\mu$m $\times$ 60 $\mu$m $\times$ 10 $\mu$m mobile microrobot platform introduced in~\cite{donald_untethered_2006}. A MicroStressBot has two actuated internal degrees of freedom (DOF): an untethered scratch-drive actuator (USDA) that provides forward motion, and a steering-arm actuator that determines whether the robot moves in a straight line or rotates. A single MicroStressBot can have its arm either raised or lowered, depending on the voltage applied across a substrate formed by an electrode array.  When high enough voltage is applied to the substrate, the arm is pulled into contact with the substrate and the robot rotates around the contact point. In contrast, when the voltage is reduced below a threshold, the arm is raised and the robot moves forward.

MicroStressBot control has been successfully implemented in~\cite{donald2013planning}. It has been shown that if the pull-down and release voltages of robots are different, they can be independently controlled. However, it is difficult to consistently manufacture the robots so they respond in the desired way. As a result, the approach scales poorly.

\subsection{On-board Physical FSM Robots}

One solution to dramatically improve the scalability of the micro-robot swarm control is to make the robots respond to a temporal sequence of (a small number of) voltage levels rather than to the voltage directly. Finite State Machines (FSMs) can accept a set of input sequences~\cite{sipser1996introduction} (sequences of control signal levels). Previously, \cite{paprotny2017finite} proposed on-board MEMS Physical FSM (PFSM) that upon the acceptance of a unique control signal sequence causes the behavioral change of a microrobot; they can be constructed from several basic modules that are combined together and thus fabricated efficiently. In this work, we build on this idea to propose \emph{group-based control}. In particular, we use the fact that several PFSM modules, each corresponding to one group, can be combined together so each robot can belong to several groups. 

\subsection{Dynamics}

In order to describe the group-based control, we start with a dynamic model of a MicroStressBot. The robot can freely move on a horizontal plane, so its configuration space is $SE(2)$. We will describe the state of the robot $i$ with a vector $[x_i, y_i, \theta_i]^T$, where $x_i$ and $y_i$ are the Cartesian coordinates of the pivot point of the robot, and $\theta_i$ is the robot orientation. The equations of motion are given by:
\begin{equation}
    \frac{d}{dt} 
    \begin{bmatrix}
    x_i\\
    y_i\\
    \theta_i
    \end{bmatrix}
  = a_i \cdot 
  \begin{bmatrix}
    \cos\theta_i\\
    \sin\theta_i\\
    0
    \end{bmatrix}\cdot u
    +(1-a_i)\cdot 
    \begin{bmatrix}
    0\\
    0\\
    1
    \end{bmatrix}\cdot \omega,
    \label{eq:continuous}
\end{equation}
where $a_i \in \{0,1\}$ is the switching input that determines whether the robot is moving forward or turning in place, while $u \in \mathbb{R}^+$ and $\omega \in \mathbb{R}^+$ are the rates of forward motion and rotation, respectively, where $\omega = u/r_i$ with $r_i$ being the turning radius of each robot. If the control inputs are piecewise constant over each epoch $\Delta T$, we obtain the following discrete-time model:
\begin{equation}
    \begin{bmatrix}
    x_i(k+1)\\
    y_i(k+1)\\
    \theta_i(k+1)
    \end{bmatrix}
    =
    \begin{bmatrix}
    x_i(k)\\
    y_i(k)\\
    \theta_i(k)
    \end{bmatrix}+
    \begin{bmatrix}
    a_{i}(k)  \cos\theta_i(k)\\
    a_{i}(k)  \sin\theta_i(k)\\
    (1-a_{i}(k))  1/r_i
    \end{bmatrix} u(k) \cdot \Delta T.
    \label{eq:discrete}
\end{equation}

Note that $u$ is the unilateral control input. In other words, Microstressbot can not go backward or turn counter-clockwise.


\section{Group-based Control}

\subsection{Group Allocation}\label{group_allo}
A common way to control multiple robots independently with a single global input is to design robots to be physically distinct so that robots can behave differently according to the same input. However, it is difficult and time-consuming to do so when the number of robots is large.
In this section, we describe an alternative method, group-based control, where robots are assigned to different (overlapping) groups. The idea is that all the robots within a group can be activated together through an on-board PFSM and move in the same way. By assigning each robot to several different groups but with none of the robots belonging to the same subset of groups, we can differentiate between the robots. Table \ref{Tab:Groups} shows how six robots can be assigned to different groups. In order to make the selection of groups unique to each robot, if we have $n$ robots we need $m=\log_2 (n+2)$ groups ($n+2$ rather than $n$ because each robot needs to belong to at least one group, and no robot can belong to all groups). For example, for six robots we need 3 groups. We start by assigning each robot a unique bit pattern, as shown in Table \ref{Tab:Groups} for 3 groups. Next, each bit position is assigned to one group (labelled with $G_i$, $i=1, 2, 3$). Each robot then belongs to the groups where the corresponding bit equals $1$. For example, robot $R_1$ belongs to group $G_1$. Instead, robot $R_3$ belongs to groups $G_2$ and $G_3$, etc. It is clear that this method guarantees that each robot has a distinct group allocation.


\begin{table}[htb]
\vspace{-2mm}
\begin{center}
 \begin{tabular}{||c | c  c  c c c c||} 
 \hline
 & \multicolumn{6}{c ||}{Robot} \\
 Group & $R_1$ & $R_2$ & $R_3$ & $R_4$ & $R_5$ & $R_6$\\ [0.5ex] 
 \hline
 $G_1$ & 0 & 0 & 0 & 1 & 1 & 1 \\ 
 \hline
 $G_2$ & 0 & 1 & 1 & 0 & 0 & 1 \\
 \hline
 $G_3$ & 1 & 0 & 1 & 0 & 1 & 0\\

 \hline
\end{tabular}
\end{center}
\vspace{-2mm}
\caption{Allocating 6 robots to 3 groups.}
\label{Tab:Groups}
\vspace{-3mm}
\end{table}

The group allocation can be realized by the on-board PFSM. At each time step $k$, only one group of robots is activated to go forward, the remaining robots rotate. We call a sequence of group selections an \emph{activation sequence}.
If $m$ is the number of groups, we would need $k = O(\log_2 m)=O(\log_2(\log_2(n+2)))$ bits to select a group through a PFSM\cite{paprotny2017finite}. Thus, the PFSMs for group-based control can be significantly simpler than in the case when each robot needs to be selected individually.

\subsection{Position accessibility}
We return to the switched system in Eq. \eqref{eq:continuous}. Let $q(t) = [x_1, y_1, x_2, y_2, \ldots, \theta_1,\theta_2 \ldots]$ be the state of the swarm, where $q(t) \in \mathbb{R} ^{3n}$. We can see that $q(t)[1:2 n]$ are the \emph{position states}, with $q(t)[2j-1:2j]$ representing the position of robot $j$. Also, $q(t)[2n+1:3n]$ are the \emph{orientation states}, where $x(t)[2n+j]$ is the orientation of the robot $j$. We assume that the robots are identical so that the robot turning radius $r_j = r$ is the same for all robots. Next, let $\alpha_i=[ \alpha_{i, 1}, \ldots, \alpha_{i, n} ]^T, i=1,\ldots,m$ be the activation vector corresponding to group $i$ being activated. In other words, $\alpha_{i, j}=1$ if robot $j$ belongs to group $i$, and $0$ otherwise. The overall swarm dynamics can then be written as:
\begin{equation}
    \label{eqn:SS}
    \dot q(t) = f_{\nu(t)}(q(t)) \cdot u(t)
\end{equation}
where $\nu(t) \in \{1,\ldots,m\}$, and for each $i$, $f_i$ is obtained by choosing $a_j=\alpha_{i, j}$ in Eq. \eqref{eq:continuous}. This equation describes a switched driftless control-affine system~\cite{liberzon2003switching, bullo_geometric_04}.

The \emph{embedded system}~\cite{bengea2005optimal} corresponding to \eqref{eqn:SS} is defined as:
\begin{equation}
    \dot q(t) = \sum_{i = 1}^m v_{i}(t)f_{i}(q(t)) \cdot u_{i}(t)
\end{equation}
where $\sum_{i = 1}^m v_{i} =1$ and $v_{i} \in [0,1]$. By introducing $\mu_{i} = v_{i} \cdot u_{i}$, we have:
\begin{align}
    \dot q(t) = \sum_{i = 1}^m f_{i}(q(t)) \cdot \mu_{i}.
\label{eq:embedded}
\end{align}
Note that $\mu_i \ge 0$.

The new system \eqref{eq:embedded} is again in the driftless control affine form and $\mu_{i}$ are unilateral control inputs. It can be shown~\cite{bengea2005optimal} that the set of trajectories of the switched system is dense in the set of trajectories of the embedded system. This implies that any embedded system trajectory can be approximated by a switched system trajectory arbitrarily close so the accessibility and controllability of the switched system can be determined by analyzing the embedded system. 


All vector fields in \eqref{eq:embedded} obey specific patterns, which will make it possible to prove the accessibility of the system. For $m$ groups with $n = 2^m -2$ robots, $f \in \mathbb{R}^{3 n}$.  Each $f_{i}$ has the form $[\gamma_i, \beta_i]$, where $\gamma_i \in \mathbb{R}^{2n}$   and $\beta_i \in \mathbb{R}^{n}$.  Each $\beta_i$ has an alternating pattern of 1's and 0's (with the first and last pattern truncated), with each pattern having length $2^{m-i}$. Further, $\gamma_i$ has an alternating pattern of $[0,0]$ and $[\cos \theta_j, \sin \theta_j]$ (with the first and last patterns truncated), with each pattern having length $2^{m-i+1}$. Formally:

\begin{align*}
    \gamma_1 &= [\mathbf{0_{2^m-2}} , \cos \theta_{2^{m-1}}, \sin \theta_{2^{m-1}}, 
     \cos \theta_{2^{m-1}+1},
     \\ & \sin \theta_{2^{m-1}+1}, \ldots, \cos \theta_{2^m-2}, \sin \theta_{2^m-2} ]^T\\
     \gamma_2 &= [\mathbf{0_{2^{m-1}-2}} , \cos \theta_{2^{m-2}}, \sin \theta_{2^{m-2}}, \ldots, \mathbf{0_{2^{m-1}}} ,
     \\ &  \cos \theta_{3\cdot 2^{m-2}}, \sin \theta_{3\cdot2^{m-2}}  ,\ldots,\cos \theta_{2^m-2}, \sin \theta_{2^m-2} ]^T\\
     \gamma_3 &= [\mathbf{0_{2^{m-2}-2}} , \cos \theta_{2^{m-3}}, \sin \theta_{2^{m-3}}, \ldots, \mathbf{0_{2^{m-2}}} ,
     \\ &   \cos \theta_{3\cdot 2^{m-3}}, \sin \theta_{3\cdot 2^{m-3}}, \ldots, \mathbf{0_{2^{m-2}}},  \cos \theta_{5\cdot 2^{m-3}}, \\
    & \sin \theta_{5\cdot 2^{m-3}}, \ldots, \cos \theta_{2^m-2}, \sin \theta_{2^m-2}]^T\\
    & \ldots\\
    \gamma_m &= [\cos \theta_{1}, \sin \theta_{1}, \mathbf{0_{2}}, \cos \theta_{3}, \sin \theta_{3}, \ldots ]^T\\[2mm]
    \beta_1 &=[ \mathbf{1_{2^{m-1}-1}}, \mathbf{0_{2^{m-1}-1}}]^T \cdot 1/r \\
    \beta_2 &=[ \mathbf{1_{2^{m-2}-1}},\mathbf{0_{2^{m-2}}},\mathbf{1_{2^{m-2}}},\mathbf{0_{2^{m-2}-1}}]^T \cdot 1/r \\
    \beta_3 &=[\mathbf{1_{2^{m-3}-1}},\mathbf{0_{2^{m-3}}},\mathbf{1_{2^{m-3}}},\mathbf{0_{2^{m-3}}},\\ &\mathbf{1_{2^{m-3}}},\mathbf{0_{2^{m-3}}},\mathbf{1_{2^{m-3}}},\mathbf{0_{2^{m-3}-1}}]^T
    \\ & \ldots \\
    \beta_m &= [ 0,\ldots, 1, 0, 1, 0, \ldots, 1]^T \cdot 1/r
  \end{align*}


Consider the example with 3 groups and 6 robots (see Table \ref{Tab:Groups}). Control vector fields $f_i \in \mathbb{R}^{18}$ with position states $[p_1, \ldots, p_6]^T \in \mathbb{R}^{12}$ and orientation states $[\theta_1,\ldots, \theta_6]^T \in \mathbb{R}^6$ are  shown in the Tables \ref{Tab:3g6rf} and \ref{Tab:3g6rh}. Vector fields $\beta_1, \beta_2, \beta_3$ have the pattern lengths $4,2,1$ respectively, with the first and last $0$ and $1$ being truncated. Similarly, vector fields $\gamma_1, \gamma_2, \gamma_3$ have the pattern lengths $8,4,2$ respectively, with the first and last pair $[0,0]$ and $[\cos, \sin]$ being truncated. Note that $\gamma_i$ mirrors the group allocation pattern in Table \ref{Tab:3g6rf} while $\beta_i$ corresponds to the negation of the allocation pattern.  The truncations result from the columns corresponding to all 0s (first) and all 1s (last) missing in Table  \ref{Tab:Groups}.

\begin{table}[htb]
\vspace{-2mm}
\begin{center}
 \begin{tabular}{|| c | c c c c c c ||} 
  \hline
& \multicolumn{6}{c ||}{Position states $\gamma$} \\
  &  $p_1$& $p_2$& $p_3$& $p_4$& $p_5$ & $p_6$ \\ [0.5ex] 
 \hline\hline
 $\gamma_1$  & 0& 0& 0& $cs_4$& $cs_5$&$cs_6$ \\
 \hline
  $\gamma_2$  & 0& $cs_2$& $cs_3$& 0& 0&$cs_6$ \\
 \hline
  $\gamma_3$  & $cs_1$& 0& $cs_3$& 0& $cs_5$&0 \\
 \hline
\end{tabular}
\end{center}
\vspace{-2mm}
\caption{The pattern of the position vector fields $f_1$, $f_2$ and $f_3$; $p_i=cs_i$ means that $i$th robot position state is $[\cos \theta_i, \sin \theta_i]^T$, while $p_i=0$ corresponds to $[0,0]^T$.}
\label{Tab:3g6rf}
\vspace{-3mm}
\end{table}

\begin{table}[htb]
\vspace{-1mm}
\begin{center}
 \begin{tabular}{|| c | c c c c c  c||} 
  \hline
  & \multicolumn{6}{c ||}{Orientation states $\beta$ } \\
  & $\theta_1$& $\theta_2$& $\theta_3$& $\theta_4$& $\theta_5$ &$\theta_6$ \\ [0.5ex] 
 \hline\hline
 $\beta_1$ & 1& 1& 1& 0& 0&0 \\
 \hline
 $\beta_2$  & 1& 0& 0& 1& 1&0 \\
 \hline
  $\beta_3$  & 0& 1& 0& 1& 0&1 \\
 \hline
\end{tabular}
\end{center}
\vspace{-2mm}
\caption{The pattern of the orientation states in vector fields $f_1$, $f_2$, and $f_3$; $\theta_i = 1$ means the orientation state of robot $i$ is 1/r.}
\label{Tab:3g6rh}
\vspace{-3mm}
\end{table}

We next recall some definitions. Assume a general driftless control-affine system:\vspace{-2mm}
\begin{equation}\label{eq:controlaffine}
    \dot{x}(t)=\sum_{i=1}^{m} f_{i}(x(t)) u_{i}(t).
\end{equation}
\vspace{-2mm}

\begin{definition}
The \emph{reachable set} $\mathcal{R}_{p}(T)$ of system \eqref{eq:controlaffine} at time $T \geq 0$, subject to the initial condition $x(0)=p$ is the set $\mathcal{R}_{p}(T)=\{x(T, u): x(0)=p \text { and } u:[0, T] \mapsto U \subseteq \mathbb{R}\}$.
\end{definition}

\begin{definition}
    The system \eqref{eq:controlaffine} is \emph{accessible} from $x(0)=p$, if the reachable sets $\mathcal{R}_{p}(t)$ has non-empty interior for all $t>0$.
\end{definition}

  If $\mathcal{F}$ is a set of control vector fields, the Lie algebra, spanned by all iterated Lie brackets of the vector fields $f_{i}$, is denoted as $L(\mathcal{F})$.

\begin{lemma}[Accessibility \cite{sussmann1987general, kawski_combinatorics_2002}] The affine system \eqref{eq:controlaffine} initialized at starting state $x(0)=0$ is accessible if and only if $\operatorname{dim} L\left(f_{1}, \ldots, f_{m}\right)(0)=n$
\end{lemma}

This is usually called Lie algebra Rank Condition. Note that control vector field $f_i$ does not need to be symmetric for accessibility \cite{lynch1999controllability}. In other words, $-f_i$ does not need to be in the set $\mathcal{F}$. The fact that for our system the control is unilateral will therefore not affect accessibility.
\begin{theorem}\label{therom acc}
    The system \eqref{eq:embedded} is accessible for all position states $p$ in $\mathbb{R}^{2 n}$, in the sense of position states. Mathematically, $\operatorname{dim} L\left({f}_{1},  \ldots, {f}_{m}\right)(x)[1:2n]= 2n$.
\end{theorem}

We next outline the sketch of the proof of Theorem \ref{therom acc}. The idea is that we can find a pattern of nested brackets which makes Lie algebra  full rank. Let $S^{(1,3,\mathbf{\hat{2}}_{m-2})} (\mathcal{F})$ be a subspace of $L(\mathcal{F})$, spanned by all brackets containing exactly $1$ factor of a vector field $f_i$, at most $3$ factors of another vector field $f_j$, and at most $2$ factors of any other vector field $f_l$ besides $f_i, f_j$. In other words, $S^{(1,3,\mathbf{\hat{2}}_{m-2})}(\mathcal{F})$ is a set of homogeneous components, i.e. $k_i = 1, k_j \leq 3$ and $k_l\leq 2$, $\forall{l} \neq i,j$

The key reason behind this introduced subspace $S^{(1,3,\mathbf{\hat{2}}_{m-2})}(\mathcal{F})$ is that the vector field $f_i$ (with constant $\beta_i$) in Lie bracket can eliminate some robot's position states in the resulting vector. Then, applying another layer of orientation vector field $f_i$ will switch the entries corresponding to the robot's $x, y$ position and change their signs in the resulting vector field. Thus, by nesting different combinations of $f_i$ in the Lie brackets, we can finally get the vector field of a single robot's position states. Therefore, we can prove that $S^{(1,3,\mathbf{\hat{2}}_{m-2})}(\mathcal{F})$  is full rank in terms of position states. Since $S^{(1,3,\mathbf{\hat{2}}_{m-2})}(\mathcal{F})$  is a subspace of Lie algebra $L(\mathcal{F})$, $L(\mathcal{F})$ also spans the whole space in terms of robot positions, i.e. $rank(L(\mathcal{F})[1:2N]) = 2n$. In other words, the system \eqref{eq:embedded} is locally accessible at all states.

Revisit the example of 6 robots with 3 groups. Examples of Lie brackets between the vector fields are shown in Table \ref{Tab:Liebrackets}. We use $($ad$ v,w) = [v, w]$ and denote $[v, ($ad$^{i}v, w)]$ as $($ad$^{i+1}v, w)$. We use $s_i$ and $c_i$ for $\sin \theta_i$ and $\cos\theta_i$, respectively.

\begin{table}[htb]
\begin{center}
 \begin{tabular}{|| c | c c c c ||} 
  \hline
 & \multicolumn{4}{c ||}{Position states of Robot 2-5} \\
  Lie brackets &  $p_2$& $p_3$& $p_4$& $p_5$ \\ [0.5ex] 
 \hline\hline
 
$[f_1, f_2]$  & $-s_2, c_2$& $-s_3, c_3$& $s_4, -c_4$&  $s_5, -c_5$\\

 \hline

  $($ad$^2 f_1, f_2)$ & $-c_2,-s_2$& $-c_3,-s_3$& 0& 0\\

 \hline
    $($ad$^3 f_1, f_2)$ & $s_2,-c_2$& $s_3,-c_3$& 0& 0 \\
 \hline
    $[f_3 ,($ad$^3 f_1, f_2)]$ & 0& $c_3,s_3$&  0 & 0 \\
 \hline
   $($ad$^2 f_3 ,($ad$^3 f_1, f_2))$ & 0& $-s_3,c_3$&  0 & 0 \\
 \hline
\end{tabular}
\end{center}
\vspace{-2mm}
\caption{The pattern of Lie brackets between $f_1$, $f_2$ and $f_3$. }
\label{Tab:Liebrackets}
\vspace{-2mm}
\end{table}

The vector fields in Table \ref{Tab:Liebrackets} show a subset of $S^{(1,3,\mathbf{\hat{2}}_{m-2})}(\mathcal{F})$ , where $k_1\leq 3, k_2=1, k_3\leq2$. The bracket $[f_1,f_2]$ has a certain pattern of position states for robots 2, 3, 4 and 5, and is zero elsewhere. Then, nesting $f_1, f_3$ with the Lie bracket $[f_1, f_2]$ gives lines 2 to 5 in the table. The bottom four vectors in the table provide the rank of 4. Thus, we can conclude that robots 2 and 3 are accessible in terms of position states. By calculating the nested Lie brackets for $[f_2, f_3]$, and $[f_3, f_1]$ in a similar way, we can prove that all six robots are positionally accessible.

\subsection{Global controllability}\label{sec: nume controllable}

Proving global controllability for an arbitrary size unilateral system is trickier than proving accessibility~\cite{sontag1988controllability}. Past work~\cite{goodwine1996controllability} focuses on  eliminating the effect of the "bad" Lie brackets caused by unilateral inputs; they show that unilateral inputs can be seen as drift terms and if the corresponding "bad" brackets can be offset by lower order "good" brackets the system is small-time locally controllable (STLC). However, there is still a research gap between the global controllability of unilateral input systems and STLC of the corresponding unconstrained input systems.

Given our system is accessible and the corresponding unconstrained input system is STLC (according to~\cite{sussmann1987general, kawski_high-order_1990}), we speculate that the system is globally controllable. For systems with unilateral controls that are forced to move "forwards", the challenge for controllability is to show that the system can eventually move "backward". Intuitively, this can be achieved by making the system follow a trajectory that brings it back into the vicinity of the initial configuration. Position accessibility guarantees that in terms of positions, the swarm can move forward in every direction. We also know that since no two robots belong to the same subset of groups, their relative orientation can be changed. Together, the two observations suggest that the swarm can in fact approximately follow a closed trajectory and return to a neighborhood of the initial configuration. Our extensive numerical experiments indicate that this is in fact the case and that the position states can be arbitrarily controlled; however, formal proof is left for future work.


\section{Numerical control}

\subsection{Optimization for minimum control effort}\label{sec:mineff}

We return to the discrete-time version of \eqref{eqn:SS}. 
We would like to compute the sequence of $k$ control inputs $u_0,u_1,\ldots,u_{k-1} \in [0,c]^k$ that bring $n$ robots from the start position $p_0$ to a given goal position $p_k$. Clearly, $k$ needs to be large enough for this to be possible. Our experiments indicated that without additional constraints, the robots undergo large motions. This motivates us to explore how to move the robots efficiently, with minimum motion effort. We thus formulate the control problem as:

\begin{equation}
\begin{aligned}
\min_{u} \quad & \sum_{i=0}^{k-1} \sum_{j=1}^{n}  \alpha_{\nu_i,j} \cdot u_i\\
\textrm{s.t.} \quad & u_i \geq 0, \quad i\in {0,\ldots,k-1}\\
    & u_i \leq c, \quad  i\in {0,\ldots,k-1}\\
  & (\sum_{i=0}^{k-1} \mathbf{f_{\nu_i}(q_i)} \cdot u_i) [1:2n]  = \mathbf{P_k}-\mathbf{P_0}
\end{aligned}
\label{eq:cvx}
\end{equation}
where  $\nu_i \in \{1,\ldots,m\}$ is the given switching signal, $\mathbf{f_{\nu_i}(\cdot)}$ is determined by the group activated at time $i$ and the state transition is given by $\mathbf{q_i} = \mathbf{q_{i-1}} + \mathbf{f_{\nu_{i-1}}(q_{i-1})}\cdot u_{i-1}$. The inputs $u = [u_0, u_1, \ldots, u_{k-1}]$ are non-negative and bounded by the constant $c$. We solve this nonlinear optimization problem using the \emph{fmincon} function in MATLAB optimization toolbox.

\section{Collision Avoidance}

To address the collision avoidance problem, one could formulate it as an optimization problem with distance constraints. But the resulting problem becomes computationally intractable as the number of robots increases. We thus explore a local collision avoidance method where we perform group-based random walk whenever future collisions are predicted. 

The strategy consists of two phases: a local collision avoidance (group-based random walk) and numerical control.  A successful path is generated through switching back and forth between the two phases and reaching the goal by the numerical control at last. We call the state after a numerical control phase and a local collision avoidance the \emph{intermediate state}. We assume there exists an intermediate state $q_{int}$ in the collision-free space $C_{free}$ so that we can navigate robots from $q_{int}$ to $q_{goal}$ without collision by applying the numerical control. 


The overall collision avoidance method is outlined in Algorithm \ref{alg:pseudo}, where $g(.)$ is the robot dynamics and $NC(.)$ is the minimum effort control described in \ref{sec:mineff}. The system switches to the random-walk phase if robots are close to other robots or obstacles. When the random-walk finishes the robots are apart from each other (or possible obstacles) for at least a distance $d$. The proof of the completeness of the resulting algorithm is similar to~\cite{carpin2005motion} and is omitted due to space constraints.

 
\begin{algorithm}
    \caption{Numerical control with Collision Avoidance (NC-CA)}
     \begin{algorithmic}[1]
    \renewcommand{\algorithmicrequire}{\textbf{Input:}}
    \renewcommand{\algorithmicensure}{\textbf{Output:}}
    \REQUIRE $q_{start}$, $q_{goal}$, number of steps $k$, robot dynamics $f(.)$, function NC(.) that computes the solution to \eqref{eq:cvx}, activation sequence $\nu$
    \ENSURE  robot path $q$
    \STATE \emph{Initialisation} : $q = q_{start}$
    \WHILE{$q \neq q_{goal}$}
        \STATE  $u \gets NC(q, q_{goal}, \nu)$;
        \FOR{$i=1:k$}
            \STATE $q \gets g(q, u, \nu)$;
            \IF{future collision from $q$}
                \STATE $q \gets$ Random Walk($q$);
                \STATE \textbf{break}
            \ENDIF
        \ENDFOR
    \ENDWHILE
    \end{algorithmic}
    \label{alg:pseudo}
\end{algorithm}



\section{Simulation Results}\label{sec:results}

Our simulations use MATLAB with mobile robotics simulation toolbox. All experiments ran on a 2.7 GHz Macbook.

\subsection{Group-based Control}
\label{sec:group}

\begin{figure}[ht]
    \centering
  \subfloat[\label{1a}]{%
      \includegraphics[width=0.48\linewidth]{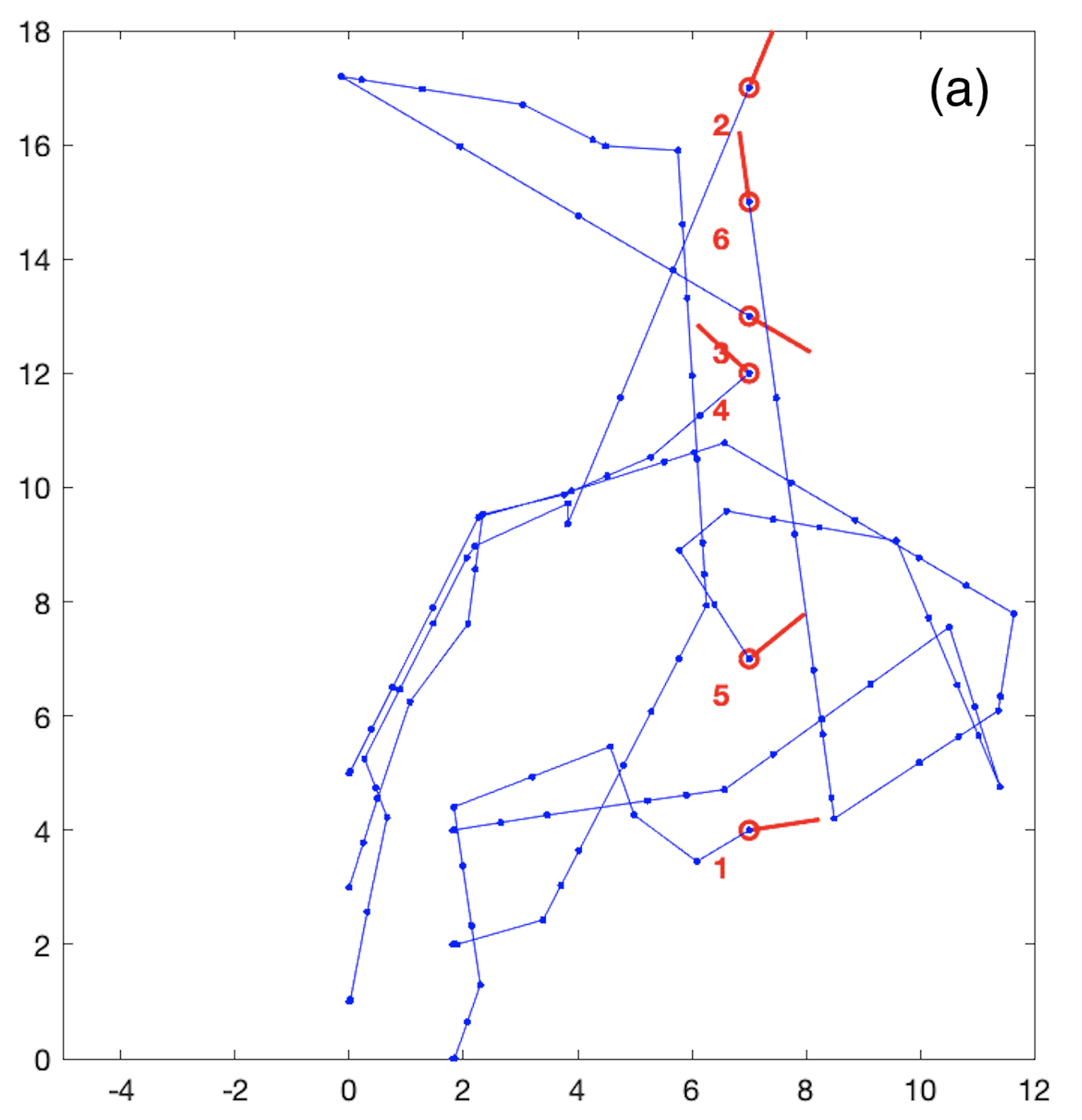}}
      \hfill
  \subfloat[\label{1b}]{%
        \includegraphics[width=0.49\linewidth]{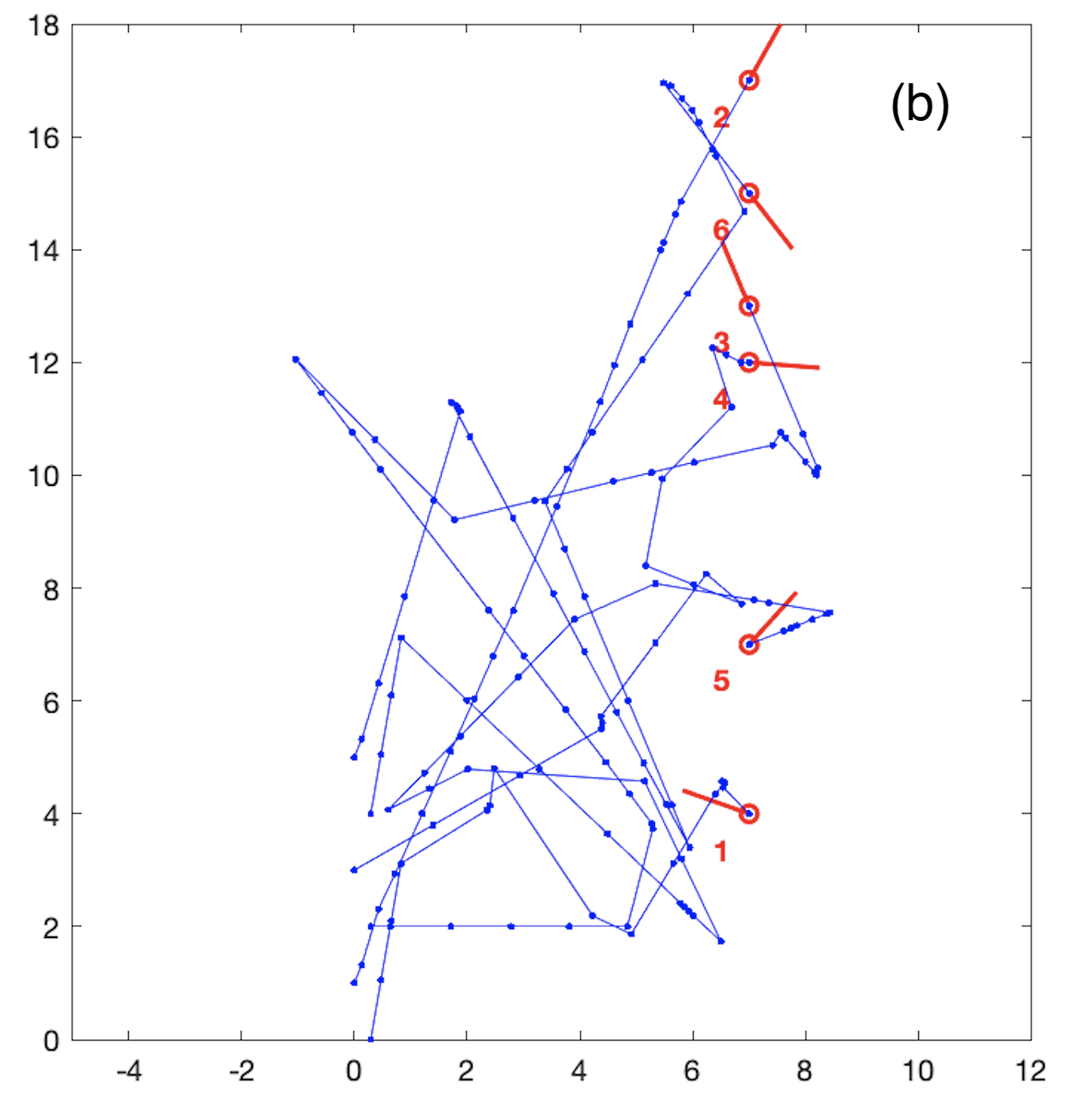}}
    \\[-5mm]
  \subfloat[\label{1c}]{%
      \includegraphics[width=0.485\linewidth]{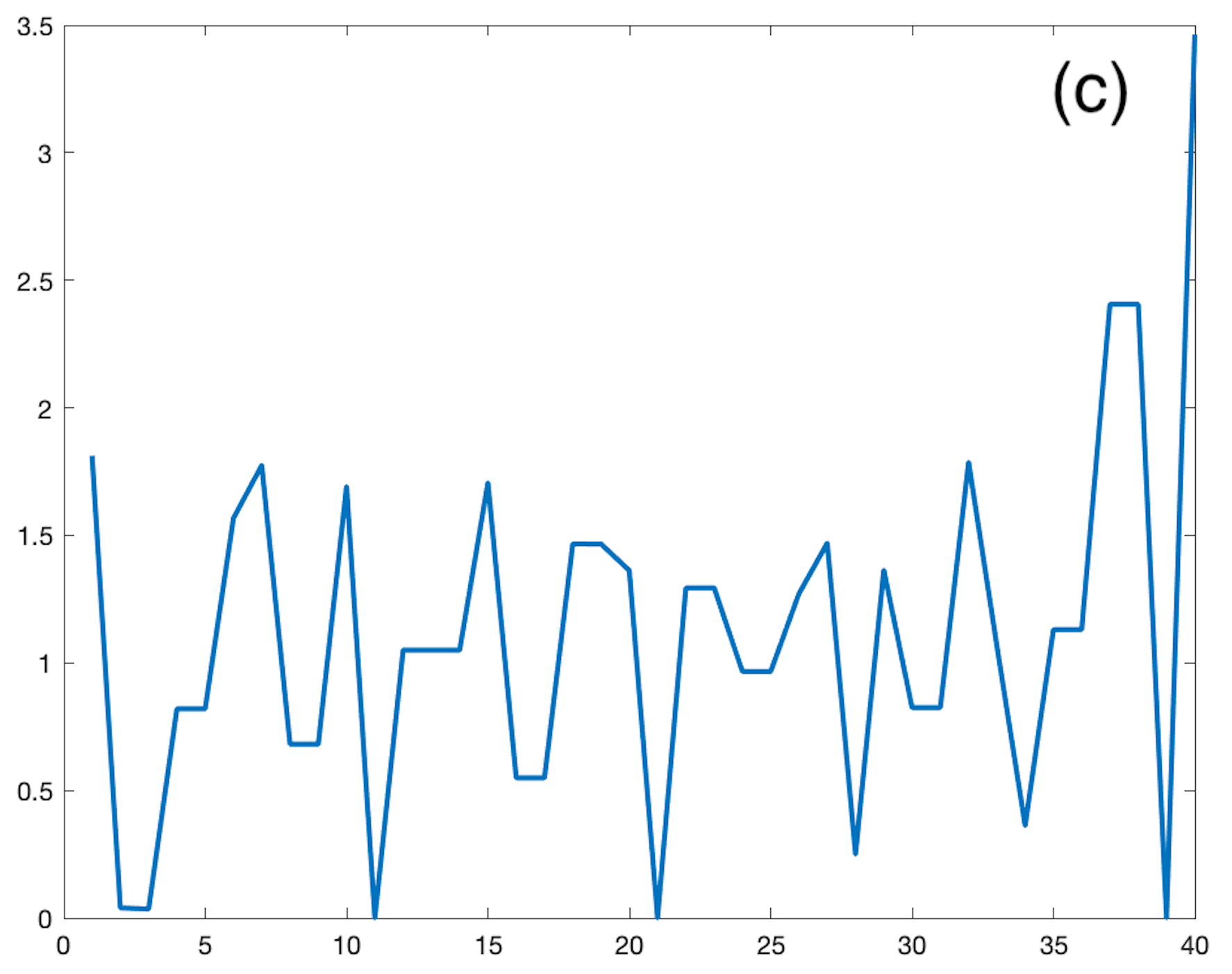}}
     \hfill
  \subfloat[\label{1d}]{%
        \includegraphics[width=0.49\linewidth]{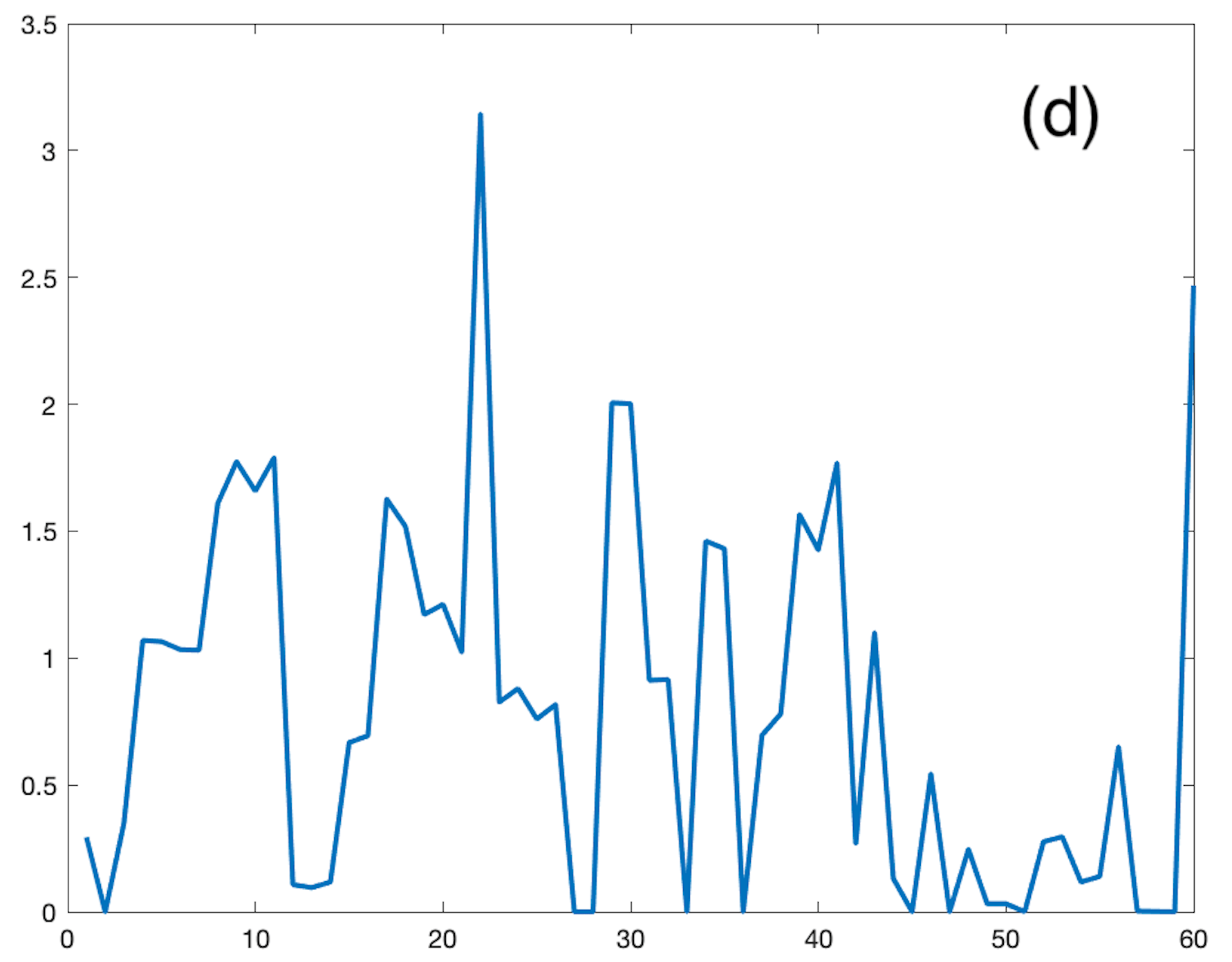}}
        \vspace{-5mm}
  \caption{(a) and (b) Minimum-effort trajectories (Sec. \ref{sec:mineff}) for $k=40$ and $60$, respectively; (c) and (d) the corresponding control inputs $u$}
  \label{fig1} 
  \vspace{-4mm}
\end{figure}

\begin{figure} [ht]
    \centering
  \subfloat[\label{2a}]{%
      \includegraphics[width=0.5\linewidth]{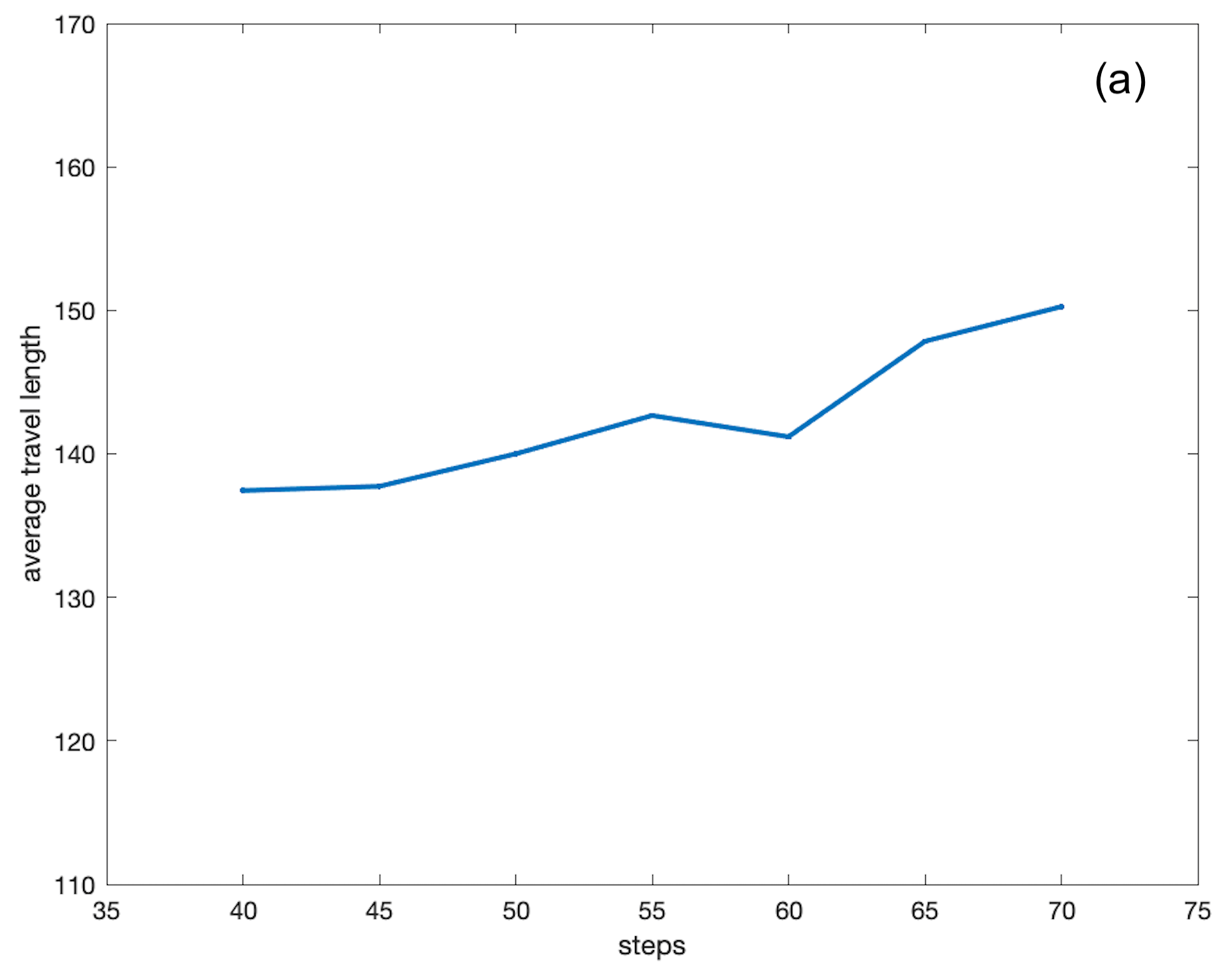}}
        \hfill
  \subfloat[\label{2c}]{%
      \includegraphics[width=0.48\linewidth]{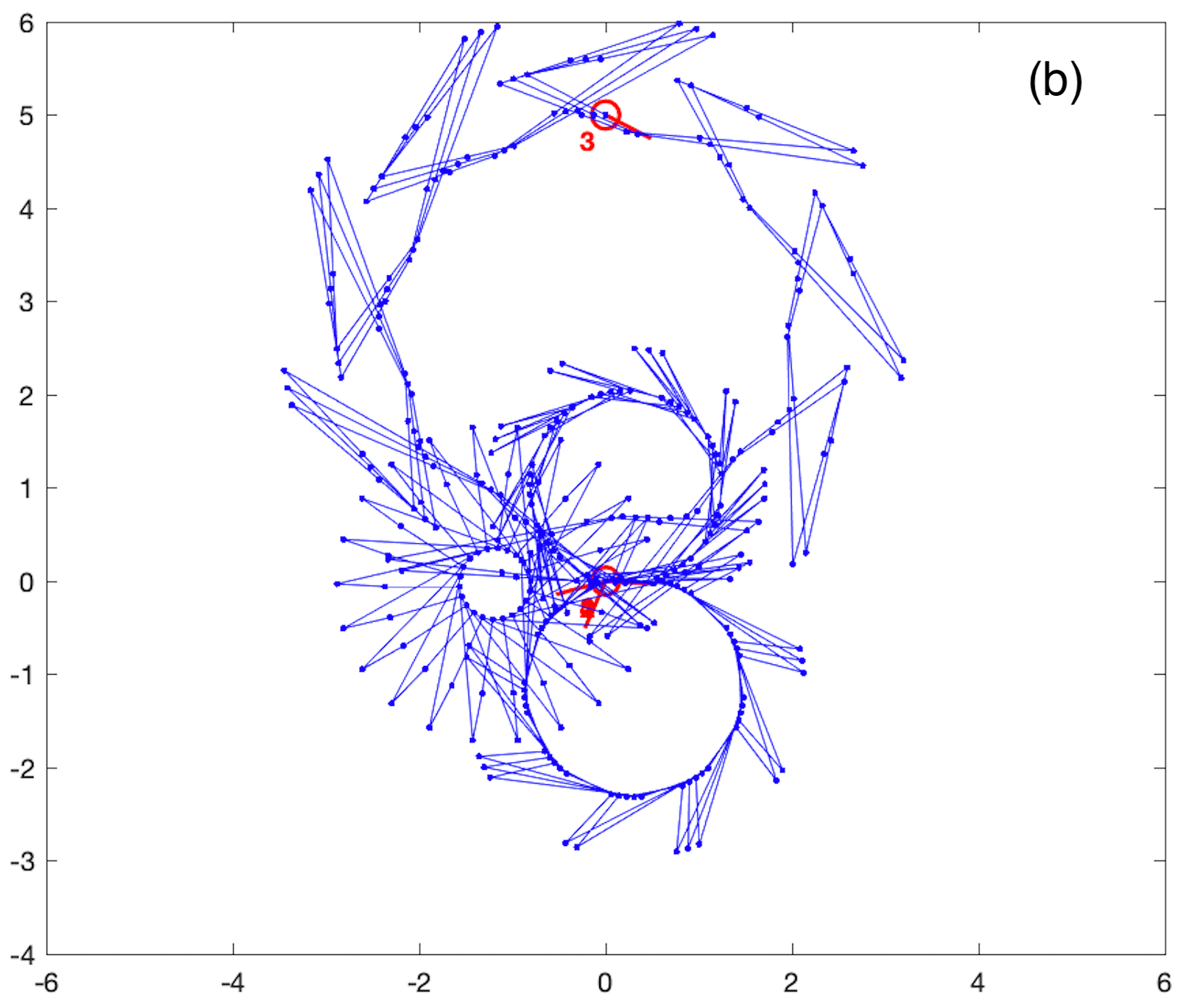}}
        \\[-5mm]
   \subfloat[\label{2b}]{%
        \includegraphics[width=0.49\linewidth]{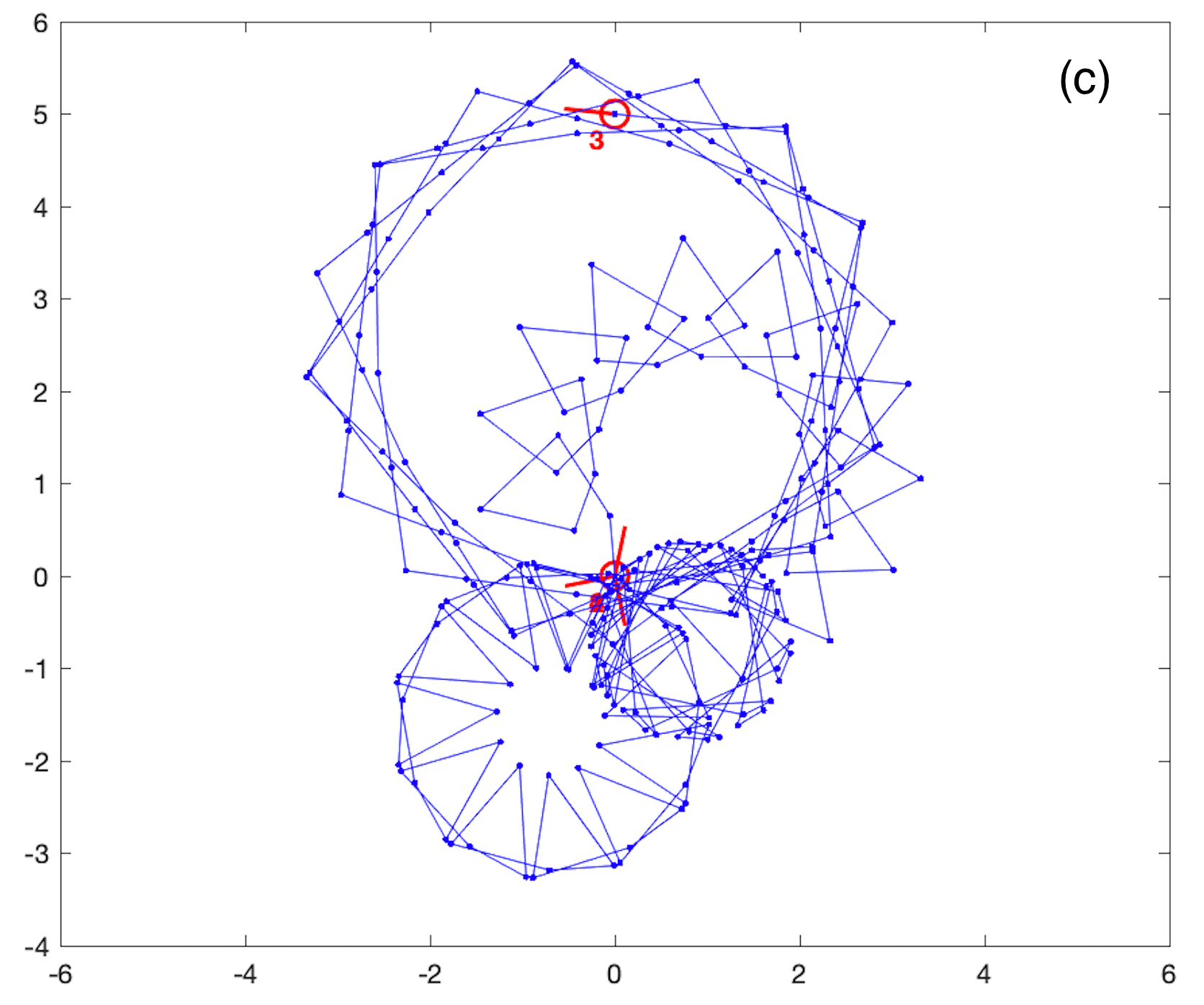}}
           \hfill
  \subfloat[\label{2d}]{%
        \includegraphics[width=0.49\linewidth]{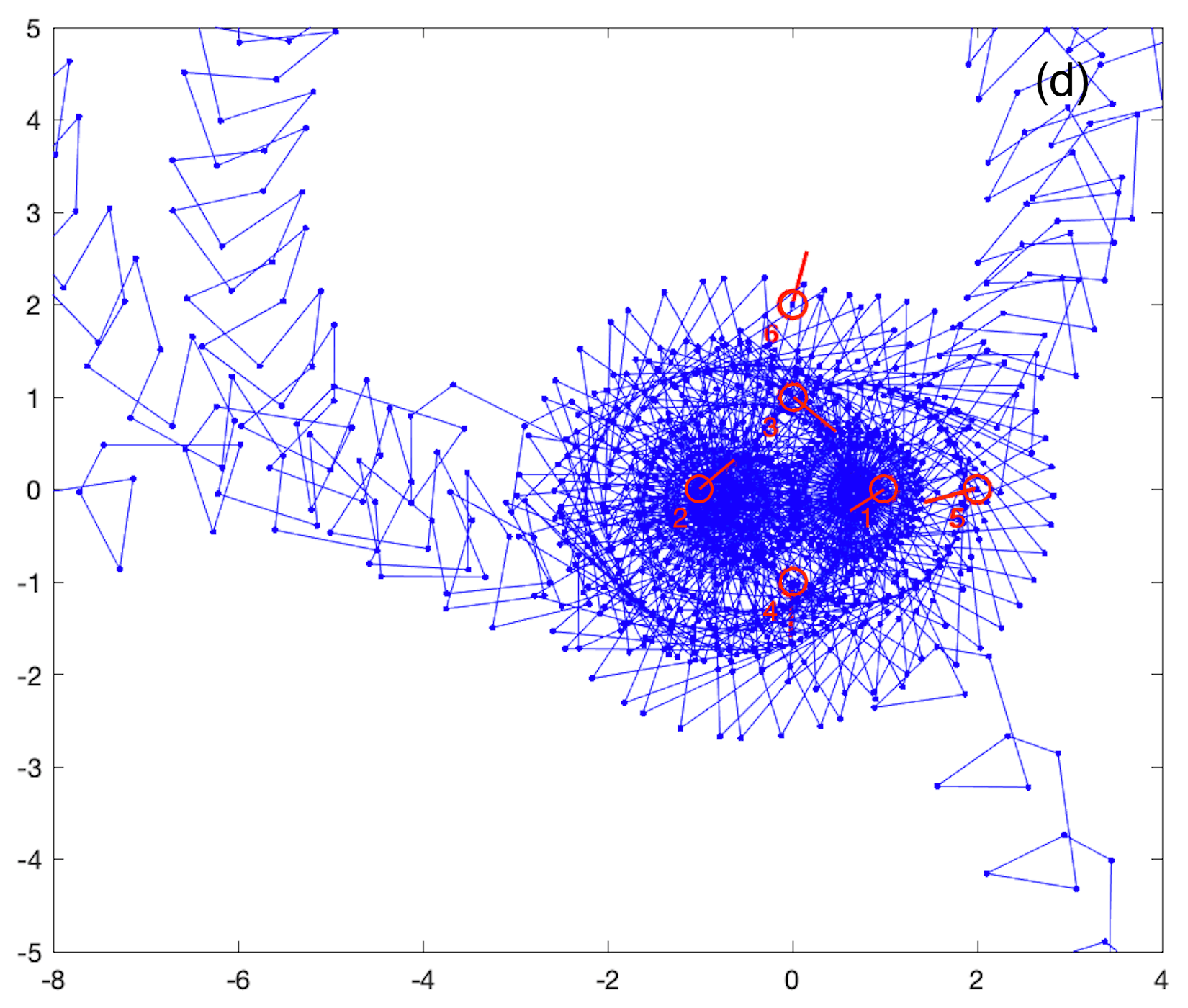}}
  \vspace{-5mm}
  \caption{(a) control steps vs. average travel length; (b), (c) and (d) robots move along the circular path}
  \label{fig2}
  \vspace{-5mm}
\end{figure}

We used a $25 \times 25$ square environment. Fixed start and goal positions for 6 robots are $[0,0;0,1;0,2;0,3;0,4;0,5]^T$ and $[7,4;7,17;7,13;7,12;7,7;7,15]^T$. All robots' initial angles were 0. Figures \ref{1a} and \ref{1b} show the results of running the fmincon minimum control effort method in Section \ref{sec:mineff} on 6 robots organized in 3 groups. All robots have the same turning radius $r = 0.05$ rad/step. The control strategy can be calculated in a matter of seconds. The total travel distances were minimized and equaled about $133.34$ and $142.79$ for (random) activation sequences of lengths $k= 40$ and $60$, respectively. However, as can be seen, in both cases robots move back and forth, leading to frequent collisions. This behavior results because the robots are restricted to only moving forward. To explore the influence of the length of the control sequence $k$ on the travel distance, we examine each sequence length 20 times (the sequence is chosen at random) and take the average; the results are shown in Figure \ref{2a}. Travel distance does not increase much if we use more steps, but computing time decreases significantly. Figures \ref{1b} and \ref{1d} show the corresponding control inputs. We can see that in both cases, inputs are constrained to lie in $[0,3.5]$ and are not sparse.


\subsection{Collision avoidance}

We compare the Local collision avoidance with Composition Rapidly Exploring Randomized Tree (RRT) \cite{hvvezda2019improved}. The sampling-based RRT motion planner searches the robots' feasible state space by building a tree data structure of possible robot motions rooted at the start position of the system. The Composition RRT we implemented attempts to grow the tree in robots' position composite space by iteratively calling a local planner until the goal is reached. The local planner samples a point in composition space and randomly selects a group to activate, then steers the tree to grow by choosing a global translation input $u$ and an angular input $w$. 

\begin{figure} [htp]
    \centering
  \subfloat[\label{3a}]{%
       \includegraphics[width=0.49\linewidth]{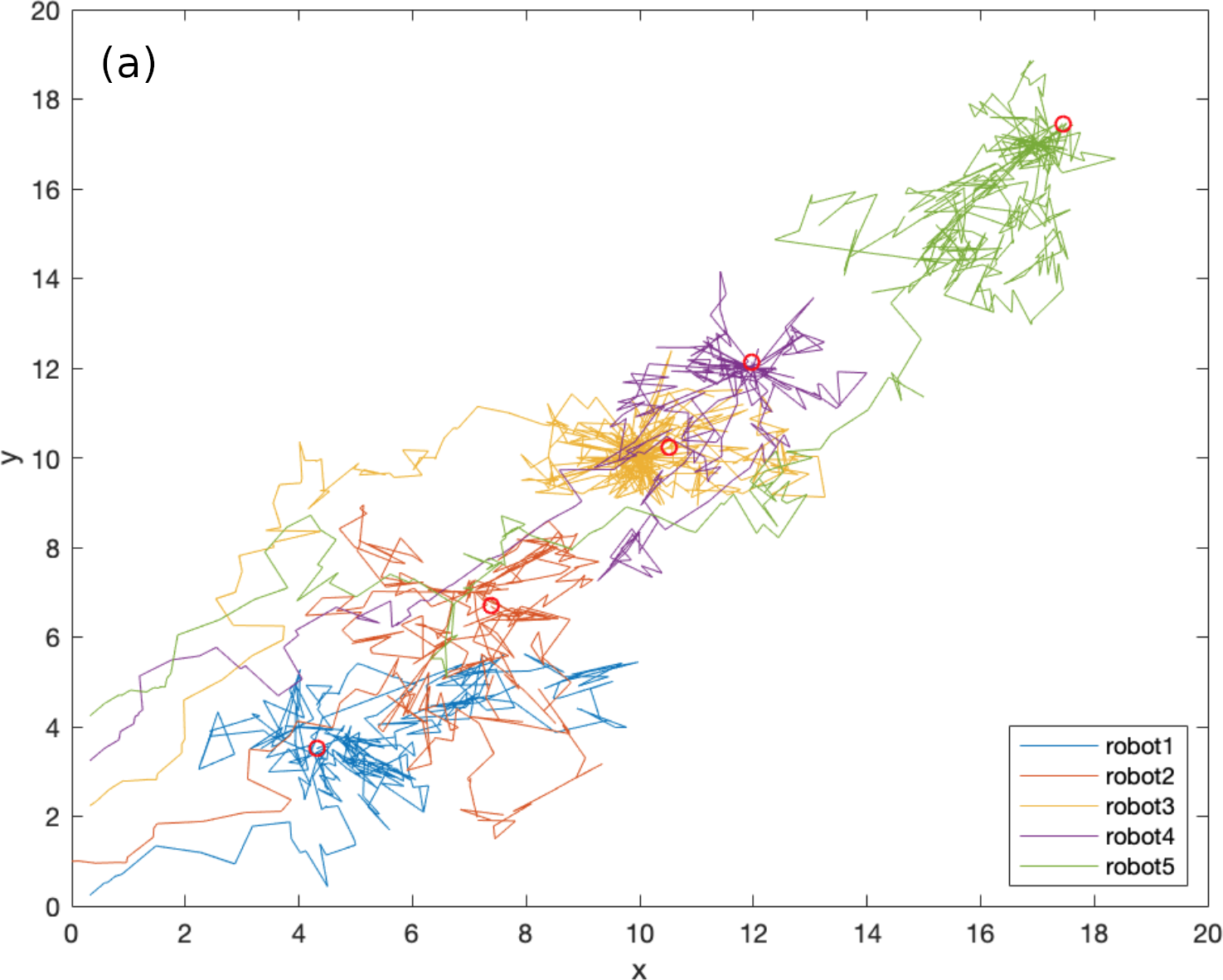}}
        \hfill
  \subfloat[\label{3b}]{%
        \includegraphics[width=0.49\linewidth]{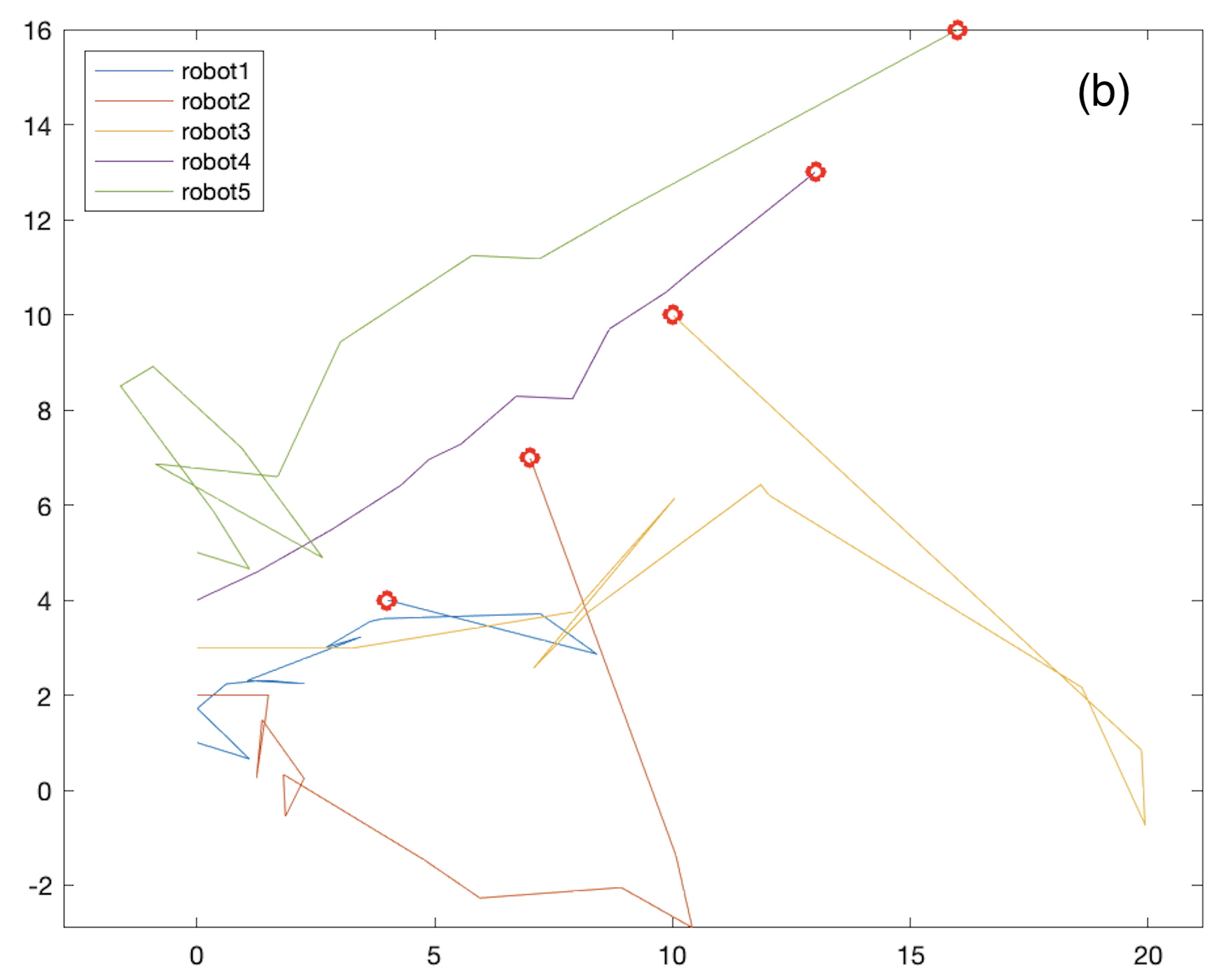}}
        \vspace{-5mm}
  \caption{(a) Collision-free trajectories obtained by composition-RRT; (b) collision-free trajectories obtained by Algorithm \ref{alg:pseudo}.}
  \label{fig3}
  \vspace{-3mm}
\end{figure}

We consider 5 robots and deploy them in a $20\times 20$ square environment. Composition space for RRT consists of five robots' position space, denoted as  $\mathcal{F}^{10}$, where $\mathcal{F}= [0,20]$.  Figure \ref{fig3} shows the collision-free path found by Composition RRT and with our proposed Algorithm \ref{alg:pseudo}. Composition RRT gives a very inefficient path. Moreover, it causes a lot of redundant paths in order to drive robots to the vicinity of the goal. Comparing the execution times, Composition RRT takes about 10 hours on average, while local collision avoidance can solve it within a minute. We can conclude that for our problem, local collision avoidance is much more efficient than a global probabilistic algorithm when handling multiple robots. 

\begin{table}[htp]
\begin{center}
 \begin{tabular}{||c c c||} 
 \hline
 Methods & Minimum affort & Algorithm \ref{alg:pseudo}\\ [0.5ex] 
 \hline\hline
 Travel length & 154.1513 & 344.7541 \\ 
 \hline
 Execution Time(s) & 9.7456 & 73.5157  \\
 \hline
 Time-out rate & 0 &  0.01 \\ [1ex]
 \hline
\end{tabular}
\end{center}
\vspace{-2mm}
\caption{Comparison of local collision avoidance and the minimum-effort approach in Sec. \ref{sec:mineff}.}
\label{tab:algcompare}
\vspace{-3mm}
\end{table}

To get a better understanding of local collision avoidance performance, we ran it on 200 random initial and final configurations, which are sampled in $[0,20]^2$ square space. Table \ref{tab:algcompare} shows the comparison between the default control method and Algorithm \ref{alg:pseudo} in terms of total traveled distance, execution time, and algorithm time-out rate. The time-out event occurs when the execution time is longer than 15 times the average. All experiments ran successfully and only $1\%$ of cases takes a long time. On average, travel length is about 2 to 3 times longer than the path with collisions. Thus, we conclude that the path is efficient and requires a reasonable control effort. Moreover, by comparing the execution time, the local path search takes less than a minute to avoid all collisions, making it a good candidate for online collision avoidance even in time-sensitive applications.

\begin{figure} [htp]
    \centering
  \subfloat[\label{4a}]{%
       \includegraphics[width=0.49\linewidth]{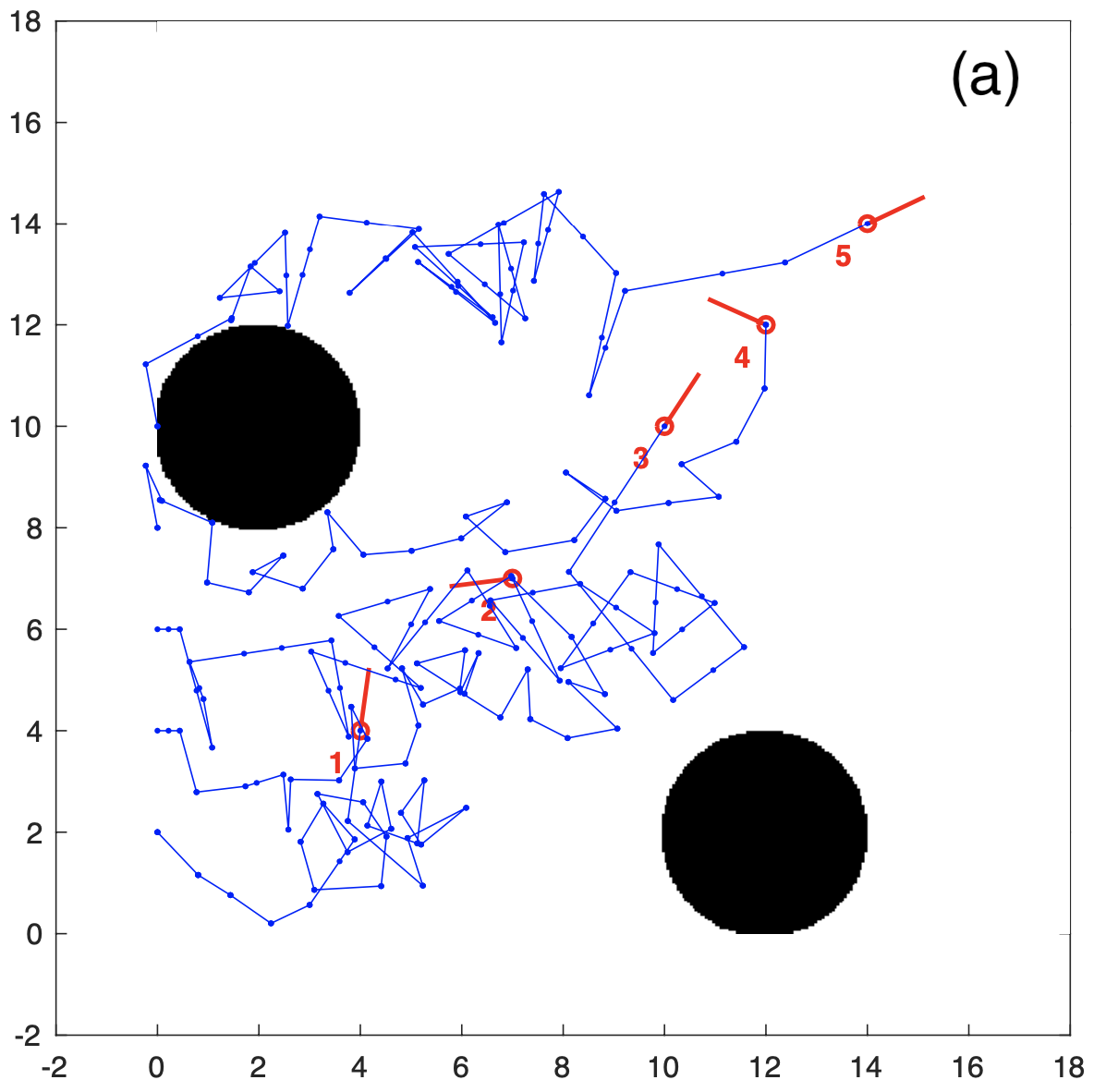}}
        \hfill
  \subfloat[\label{4b}]{%
        \includegraphics[width=0.49\linewidth]{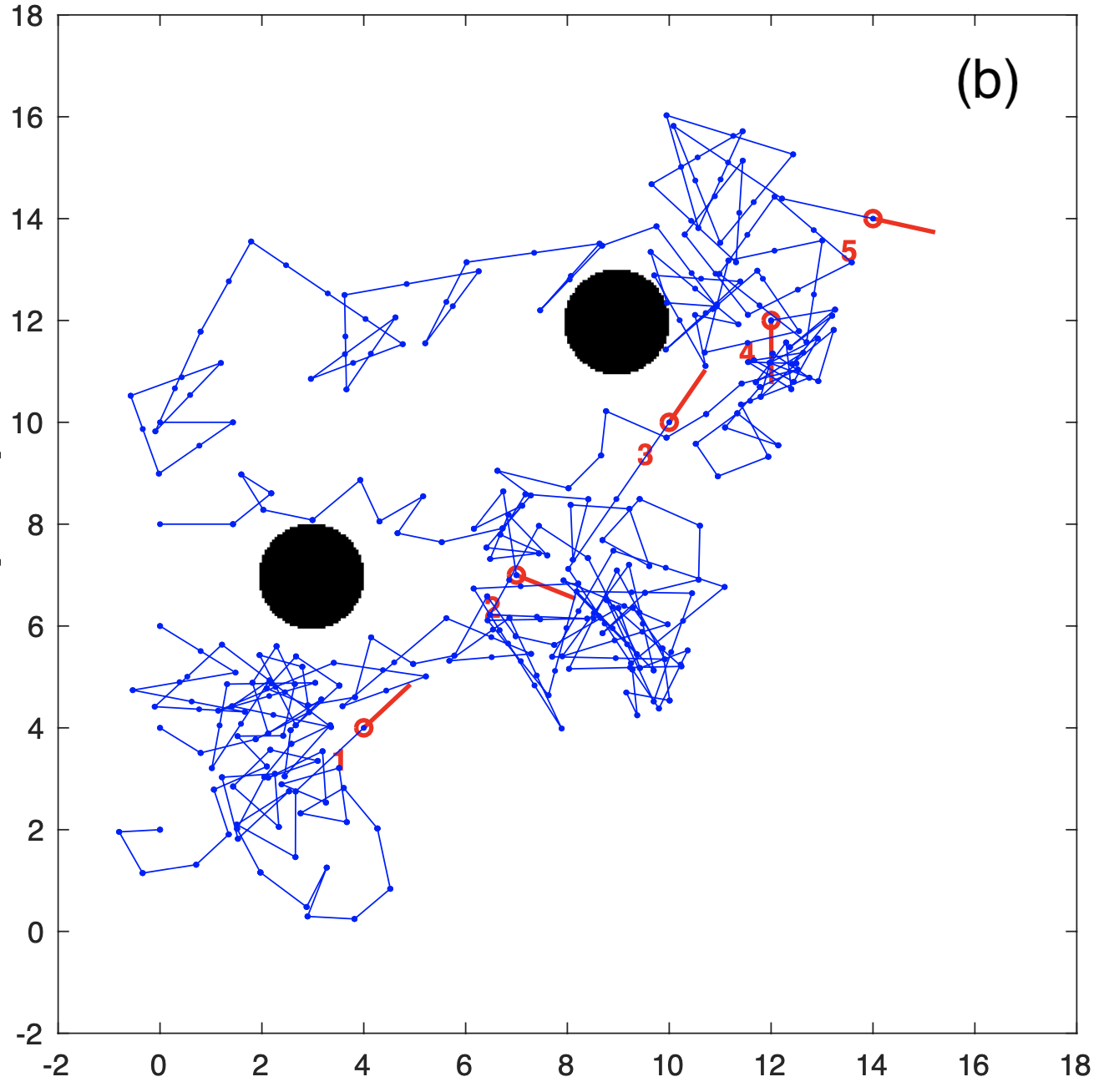}}
        \vspace{-5mm}
  \caption{(a) Obstacle avoidance trajectories in environment a; (b) Obstacle avoidance trajectories in environment b .}
  \vspace{-4mm}
  \label{fig4}
 \end{figure}

We further investigate how well the collision avoidance algorithm works in environments with obstacles. We designed two settings with two circular obstacles with radii of 2, and 1, respectively. Fig. \ref{fig4} shows a random run in each environment setting. By adding obstacles, we observe a more zig-zag pattern of robot motions than before. There are two reasons for such behavior.
First, obstacles are very close to the final positions. Specifically, we put obstacle in Fig. \ref{4b}  closer to the goal than in Fig. \ref{4a}. Even though obstacles in Fig. \ref{4a} have larger radii, Fig. \ref{4a} provides a simpler path. Second, we only have two degrees of freedom to control five robots (selecting the group and selecting the control $u$). Some robots inevitably wander around aimlessly while we move other robots away from obstacles.

\begin{table}[htp]
\begin{center}
 \begin{tabular}{||c c c||} 
 \hline
 Environment & a & b \\ [0.5ex] 
 \hline\hline
 Ave Travel length & 491.6381 & 758.9985 \\ 
 \hline
 Min Travel length & 405.3294 & 445.4527  \\
 \hline
 Time-out rate & 0 &  0.04 \\ [1ex]
 \hline
\end{tabular}
\end{center}
\vspace{-2mm}
\caption{Comparison of local obstacle avoidance in two environments }
\label{tab:envcomp}
\vspace{-4mm}
\end{table}

Our findings are summarized in Table \ref{tab:envcomp} showing that the average travel distance is about 1.5 to 3.5 longer than in the environment without obstacles. Noting that the minimum length is much shorter than the average, when the map is known we can run the algorithm several times and choose the best path; each run of the algorithm typically takes less than a minute.

\subsection{Numerical evidence for global controllability}

We demonstrate global controllability through empirical experiments with 3 groups and 6 robots. The key idea is to define a set of control primitives $p$ that can be used to construct more complex trajectories. We can observe that robots move along a circular path and eventually return to the original points as we repeat the control primitive multiple times. Figure \ref{2b} shows the trajectory of 4 robots by repeating control primitive $p = [g1,g2,g3,g1,g2,g3,g1,g2,g3] = [g1,g2,g3]^3 $ 200 times. By solving for control inputs $u \in \mathbb{R}^9$, we can make only robot 3 to move while all other robots remain at their original location. Figure \ref{2c} gives the trajectories of 4 robots for the control primitive $p = [g1,g2,g3]^4$ which repeats 150 times. Notice that robot paths in Figures \ref{2b} and \ref{2c} are similar to repeating a triangle and quadrilateral respectively. This is because the control primitives correspond to 3 and 4 repetitions of the pattern $[g1, g2, g3]$.  Figure \ref{2d} shows an example of control of 6 robots from their location $[0,0;0,1;0,2;0,3;0,4;0,5]^T$ to $x_f = [1,0;-1,0;0,1;0,-1;2,0;0,3]^T$. For all these cases, we observe that each robot will move on a path that looks circular.

\section{Conclusion}

In this paper, we introduced the novel idea of group-based control that dramatically increases the ability to control large groups of MicroStressBots with on-board PFSMs. We formally proved that the proposed method makes the positions of the robots in the swarm accessible and provided numerical evidence for global controllability. We also studied collision avoidance for the swarm and proposed a probabilistically complete algorithm for collision avoidance that can be used online. The proposed methods were extensively evaluated through simulation studies. Future work will include the formal proof of global controllability and deal with uncertainty in the operation of the robots. We also plan to evaluate the approach on a real system of MicroStressBots. Fabrication of the physical testbed is currently in progress.


\bibliographystyle{IEEEtran}
\bibliography{references}

\begin{thebibliography}{10}
\providecommand{\url}[1]{#1}
\csname url@samestyle\endcsname
\providecommand{\newblock}{\relax}
\providecommand{\bibinfo}[2]{#2}
\providecommand{\BIBentrySTDinterwordspacing}{\spaceskip=0pt\relax}
\providecommand{\BIBentryALTinterwordstretchfactor}{4}
\providecommand{\BIBentryALTinterwordspacing}{\spaceskip=\fontdimen2\font plus
\BIBentryALTinterwordstretchfactor\fontdimen3\font minus
  \fontdimen4\font\relax}
\providecommand{\BIBforeignlanguage}[2]{{%
\expandafter\ifx\csname l@#1\endcsname\relax
\typeout{** WARNING: IEEEtran.bst: No hyphenation pattern has been}%
\typeout{** loaded for the language `#1'. Using the pattern for}%
\typeout{** the default language instead.}%
\else
\language=\csname l@#1\endcsname
\fi
#2}}
\providecommand{\BIBdecl}{\relax}
\BIBdecl

\bibitem{chowdhury_controlling_2015}
S.~Chowdhury, W.~Jing, and D.~J. Cappelleri, ``Controlling multiple
  microrobots: Recent progress and future challenges,'' \emph{Journal of
  Micro-Bio Robotics}, vol.~10, no.~1, pp. 1--11, Oct. 2015.

\bibitem{li_development_2018}
J.~Li, X.~Li, T.~Luo, R.~Wang, C.~Liu, S.~Chen, D.~Li, J.~Yue, S.-h. Cheng, and
  D.~Sun, ``Development of a magnetic microrobot for carrying and delivering
  targeted cells,'' \emph{Science Robotics}, vol.~3, no.~19, Jun. 2018.

\bibitem{yu_novel_2010}
C.~Yu, J.~Kim, H.~Choi, J.~Choi, S.~Jeong, K.~Cha, J.-o. Park, and S.~Park,
  ``Novel electromagnetic actuation system for three-dimensional locomotion and
  drilling of intravascular microrobot,'' \emph{Sensors and Actuators A:
  Physical}, vol. 161, no.~1, pp. 297--304, Jun. 2010.

\bibitem{leclerc_vitro_2020}
J.~Leclerc, H.~Zhao, D.~Bao, and A.~T. Becker, ``In vitro design investigation
  of a rotating helical magnetic swimmer for combined 3-d navigation and blood
  clot removal,'' \emph{IEEE Transactions on Robotics}, vol.~36, no.~3, pp.
  975--982, Jun. 2020.

\bibitem{mahdy_ultrasound-guided_2018}
D.~Mahdy, R.~Reda, N.~Hamdi, and I.~S.~M. Khalil, ``Ultrasound-guided minimally
  invasive grinding for clearing blood clots: Promises and challenges,''
  \emph{IEEE Instrumentation Measurement Magazine}, vol.~21, no.~2, pp. 10--14,
  Apr. 2018.

\bibitem{diller_three-dimensional_2014}
E.~Diller and M.~Sitti, ``Three-dimensional programmable assembly by untethered
  magnetic robotic micro-grippers,'' \emph{Advanced Functional Materials},
  vol.~24, no.~28, pp. 4397--4404, 2014.

\bibitem{floyd_two-dimensional_2009}
S.~Floyd, C.~Pawashe, and M.~Sitti, ``Two-dimensional contact and noncontact
  micromanipulation in liquid using an untethered mobile magnetic microrobot,''
  \emph{IEEE Transactions on Robotics}, vol.~25, no.~6, pp. 1332--1342, Dec.
  2009.

\bibitem{yao_directed_2020}
T.~Yao, N.~G. Chisholm, E.~B. Steager, and K.~J. Stebe, ``Directed assembly and
  micro-manipulation of passive particles at fluid interfaces via capillarity
  using a magnetic micro-robot,'' \emph{Applied Physics Letters}, vol. 116,
  no.~4, p. 043702, Jan. 2020.

\bibitem{banerjee2012real}
A.~G. Banerjee, S.~Chowdhury, W.~Losert, and S.~K. Gupta, ``Real-time path
  planning for coordinated transport of multiple particles using optical
  tweezers,'' \emph{IEEE Transactions on automation science and engineering},
  vol.~9, no.~4, pp. 669--678, 2012.

\bibitem{pawashe2009modeling}
C.~Pawashe, S.~Floyd, and M.~Sitti, ``Modeling and experimental
  characterization of an untethered magnetic micro-robot,'' \emph{The
  International Journal of Robotics Research}, vol.~28, no.~8, pp. 1077--1094,
  2009.

\bibitem{donald_untethered_2006}
B.~R. Donald, C.~G. Levey, C.~D. McGray, I.~Paprotny, and D.~Rus, ``An
  untethered, electrostatic, globally controllable mems micro-robot,''
  \emph{Journal of Microelectromechanical Systems}, vol.~15, no.~1, pp. 1--15,
  Feb. 2006.

\bibitem{becker2014controlling}
A.~Becker, C.~Onyuksel, T.~Bretl, and J.~McLurkin, ``Controlling many
  differential-drive robots with uniform control inputs,'' \emph{The
  international journal of Robotics Research}, vol.~33, no.~13, pp. 1626--1644,
  2014.

\bibitem{paprotny_turning-rate_2013}
I.~Paprotny, C.~Levey, P.~Wright, and B.~Donald, ``Turning-rate selective
  control : A new method for independent control of stress-engineered mems
  microrobots,'' in \emph{Robotics: Science and Systems}, vol.~8, 2013, pp.
  321--328.

\bibitem{yu2018pattern}
J.~Yu, L.~Yang, and L.~Zhang, ``Pattern generation and motion control of a
  vortex-like paramagnetic nanoparticle swarm,'' \emph{The International
  Journal of Robotics Research}, vol.~37, no.~8, pp. 912--930, 2018.

\bibitem{donald2013planning}
B.~R. Donald, C.~G. Levey, I.~Paprotny, and D.~Rus, ``Planning and control for
  microassembly of structures composed of stress-engineered mems microrobots,''
  \emph{The International journal of robotics research}, vol.~32, no.~2, pp.
  218--246, 2013.

\bibitem{paprotny2017finite}
I.~Paprotny and M.~Zefran, ``Finite state machine (fms) addressable mems
  microrobots: a new paradigm for controlling large numbers of mems
  microrobots,'' in \emph{2017 International Conference on Manipulation,
  Automation and Robotics at Small Scales (MARSS)}.\hskip 1em plus 0.5em minus
  0.4em\relax IEEE, 2017, pp. 1--6.

\bibitem{kuffner_rrt-connect_2000}
J.~Kuffner and S.~LaValle, ``Rrt-connect: An efficient approach to single-query
  path planning,'' in \emph{Proceedings 2000 ICRA. Millennium Conference. IEEE
  International Conference on Robotics and Automation. Symposia Proceedings
  (Cat. No.00CH37065)}, vol.~2, Apr. 2000, pp. 995--1001 vol.2.

\bibitem{vcap2013multi}
M.~{\v{C}}{\'a}p, P.~Nov{\'a}k, J.~Vok{\v{r}}{\'\i}nek, and
  M.~P{\v{e}}chou{\v{c}}ek, ``Multi-agent rrt*: Sampling-based cooperative
  pathfinding,'' \emph{arXiv preprint arXiv:1302.2828}, 2013.

\bibitem{chiang2019rl}
H.-T.~L. Chiang, J.~Hsu, M.~Fiser, L.~Tapia, and A.~Faust, ``Rl-rrt:
  Kinodynamic motion planning via learning reachability estimators from rl
  policies,'' \emph{IEEE Robotics and Automation Letters}, vol.~4, no.~4, pp.
  4298--4305, 2019.

\bibitem{haghtalab2018provable}
N.~Haghtalab, S.~Mackenzie, A.~Procaccia, O.~Salzman, and S.~Srinivasa, ``The
  provable virtue of laziness in motion planning,'' in \emph{Proceedings of the
  International Conference on Automated Planning and Scheduling}, vol.~28,
  no.~1, 2018.

\bibitem{barraquand1991robot}
J.~Barraquand and J.-C. Latombe, ``Robot motion planning: A distributed
  representation approach,'' \emph{The International Journal of Robotics
  Research}, vol.~10, no.~6, pp. 628--649, 1991.

\bibitem{carpin2005motion}
S.~Carpin and G.~Pillonetto, ``Motion planning using adaptive random walks,''
  \emph{IEEE Transactions on Robotics}, vol.~21, no.~1, pp. 129--136, 2005.

\bibitem{sipser1996introduction}
M.~Sipser, ``Introduction to the theory of computation,'' \emph{ACM Sigact
  News}, vol.~27, no.~1, pp. 27--29, 1996.

\bibitem{liberzon2003switching}
D.~Liberzon, \emph{Switching in systems and control}.\hskip 1em plus 0.5em
  minus 0.4em\relax Springer Science \& Business Media, 2003.

\bibitem{bullo_geometric_04}
F.~Bullo and A.~D. Lewis, \emph{Geometric Control of Mechanical Systems}, ser.
  Texts in Applied Mathematics.\hskip 1em plus 0.5em minus 0.4em\relax
  Springer, 2004, vol.~49.

\bibitem{bengea2005optimal}
S.~C. Bengea and R.~A. DeCarlo, ``Optimal control of switching systems,''
  \emph{automatica}, vol.~41, no.~1, pp. 11--27, 2005.

\bibitem{sussmann1987general}
H.~J. Sussmann, ``A general theorem on local controllability,'' \emph{SIAM
  Journal on Control and Optimization}, vol.~25, no.~1, pp. 158--194, 1987.

\bibitem{kawski_combinatorics_2002}
M.~Kawski, ``The combinatorics of nonlinear controllability and noncommuting
  flows,'' International Atomic Energy Agency (IAEA), Tech. Rep. 92-95003-11-X,
  2002.

\bibitem{lynch1999controllability}
K.~M. Lynch, ``Controllability of a planar body with unilateral thrusters,''
  \emph{IEEE Transactions on Automatic Control}, vol.~44, no.~6, pp.
  1206--1211, 1999.

\bibitem{sontag1988controllability}
E.~D. Sontag, ``Controllability is harder to decide than accessibility,''
  \emph{SIAM journal on control and optimization}, vol.~26, no.~5, pp.
  1106--1118, 1988.

\bibitem{goodwine1996controllability}
B.~Goodwine and J.~Burdick, ``Controllability with unilateral control inputs,''
  in \emph{Proceedings of 35th IEEE Conference on Decision and Control},
  vol.~3.\hskip 1em plus 0.5em minus 0.4em\relax IEEE, 1996, pp. 3394--3399.

\bibitem{kawski_high-order_1990}
M.~Kawski, ``High-order small-time local controllability,'' in \emph{NonLinear
  Controllability and Optimal Control}.\hskip 1em plus 0.5em minus 0.4em\relax
  Routledge, 1990, pp. 431--467.

\bibitem{hvvezda2019improved}
J.~Hv{\v{e}}zda, M.~Kulich, and L.~P{\v{r}}eu{\v{c}}il, ``Improved discrete rrt
  for coordinated multi-robot planning,'' \emph{arXiv preprint
  arXiv:1901.07363}, 2019.

\end{thebibliography}

\end{document}